\documentclass[conference]{IEEEtran}
\IEEEoverridecommandlockouts
\usepackage{cite}
\usepackage{svg}
\usepackage{amsmath,amssymb,amsfonts}
\usepackage{algorithmic}
\usepackage{graphicx}
\usepackage{textcomp}
\usepackage{xcolor}
\def\BibTeX{{\rm B\kern-.05em{\sc i\kern-.025em b}\kern-.08em
    T\kern-.1667em\lower.7ex\hbox{E}\kern-.125emX}}
\begin{document}

\title{Turn-by-Turn Indoor Navigation for the Visually Impaired}

\author{
    \IEEEauthorblockN{Santosh Srinivasaiah}
    \IEEEauthorblockA{
        \textit{VP - Data Science and Analytics} \\
        \textit{Diaconia LLC}\\
    }
    
    \and
    
    \IEEEauthorblockN{Sai Kumar Nekkanti}
    \IEEEauthorblockA{
        \textit{Python Developer Intern - LLM Solutions} \\
        \textit{Diaconia LLC}
    }
    
    \and
    
    \IEEEauthorblockN{Rohith Reddy Nedhunuri}
    \IEEEauthorblockA{
        \textit{Python Developer Intern  - LLM Solutions} \\
        \textit{Diaconia LLC}\\
    }

}

\maketitle

\begin{abstract}
Abstract—Navigating indoor environments presents significant challenges for visually impaired individuals due to complex layouts and the absence of GPS signals. This paper introduces a novel system that provides turn-by-turn navigation inside buildings using only a smartphone equipped with a camera, leveraging multimodal models, deep learning algorithms, and large language models (LLMs). The smartphone's camera captures real-time images of the surroundings, which are then sent to a nearby Raspberry Pi capable of running on-device LLM models, multimodal models, and deep learning algorithms to detect and recognize architectural features, signage, and obstacles. The interpreted visual data is then translated into natural language instructions by an LLM running on the Raspberry Pi, which is sent back to the user, offering intuitive and context-aware guidance via audio prompts. This solution requires minimal workload on the user's device, preventing it from being overloaded and offering compatibility with all types of devices, including those incapable of running AI models. This approach enables the client to not only run advanced models but also ensure that the training data and other information do not leave the building. Preliminary evaluations demonstrate the system's effectiveness in accurately guiding users through complex indoor spaces, highlighting its potential for widespread application.
\end{abstract}


\section{Introduction}

Navigating indoor environments is a fundamental challenge for visually impaired individuals. In the United States alone, over 50 million adults experience some degree of vision loss, with nearly 4 million facing severe visual impairment even with corrective lenses, and approximately 340,000 being completely blind \cite{b1}. Unlike outdoor navigation, where technologies like GPS provide reliable guidance, indoor spaces lack universal navigation aids due to the absence of GPS signals and the complex variability of interior layouts. This limitation often results in reduced independence and increased reliance on others, impacting the quality of life for those with visual impairments.

Prior research in assistive technologies for the visually impaired can be categorized into sensory-based and vision-based systems. Sensory-based systems use non-visual cues for navigation, such as RFID-based indoor navigation systems designed for elderly and visually impaired users, which enable real-time localization and guidance \cite{b2}. Another approach utilizes Bluetooth Low Energy (BLE) beacons for indoor navigation through short-range signals \cite{b3}. Vision-based systems, on the other hand, focus on image detection and augmented reality, such as mobile-based systems that use image detection and voice translation for object recognition \cite{b4}, or augmented reality solutions like ARIANNA+, which enhance navigation using virtual paths and CNN-based object recognition \cite{b5}.

However, existing solutions frequently require specialized equipment or infrastructure modifications, such as installing beacons, RFID tags, or tactile paving. These approaches can be cost-prohibitive and are not universally scalable or practical for all indoor environments. Moreover, wearable devices designed for navigation assistance may be uncomfortable or stigmatizing for users. While sensory-based models face limitations in object detection, vision-based systems often rely on cloud-based models, raising concerns about privacy and internet dependency.

Advancements in computer vision and natural language processing, particularly the development of large language models (LLMs) and multimodal models, offer new avenues to address these challenges. Smartphones, ubiquitous in modern society, are equipped with powerful cameras and computational capabilities that can be harnessed without the need for additional hardware. Multimodal and Deep Learning models can process real-time visual data to interpret surroundings, while LLMs can generate coherent and context-aware instructions based on that data.

In this paper, we propose a novel system that leverages a smartphone's camera, state-of-the-art deep learning algorithms, multimodal models, and LLMs to provide turn-by-turn indoor navigation assistance for visually impaired users. To address privacy and connectivity concerns, our solution runs convolutional neural network (CNN) models directly on a Raspberry Pi, eliminating the need for internet access. Additionally, we implement local LLMs on the Raspberry Pi to generate descriptive and user-friendly text, ensuring all processing occurs locally and securely on the device itself. The system captures images of the user's surroundings and identifies key navigational cues such as doors, corridors, signage, and obstacles. This visual information is then translated into natural language directions delivered via audio prompts, offering a cost-effective and accessible solution for indoor navigation.
\section{Background}

\subsection{LLMs} 
Large language models have revolutionized the state of the art in understanding and generating natural languages, which is an ability acquired through training on large amounts of data to recognize patterns and allow the model to generate text that is just as human-like. Trained on diverse datasets, these models can decode complex instructions and create contextually acceptable responses. In the context of indoor navigation for visually impaired individuals, this technology will be transforming visual data into clear, real-time instructions to lead users through complicated settings. Because they can process language subtly and understand context, the instructions given are not only correct but intuitive and user-friendly. Recent work, such as what has been presented by the framework proposed by L3MVN, has demonstrated how easily LLMs are put to work in common-sense knowledge for tasks like visual target navigation, further enhancing their practical applications into real-world settings\cite{b6}.

\subsection{Raspberry PI}
Raspberry Pi serves as an economically viable microcomputer in our system, known for its portability, power efficiency, and broad applications in IoT. By hosting LLMs, multimodal models, and deep learning algorithms, it effectively offloads the computational burden from the user's smartphone, ensuring smooth and efficient operation. This edge computing setup enables local processing, maintaining data privacy by keeping sensitive information like images and navigation logs within the premises. Research has demonstrated Raspberry Pi's effectiveness in edge-based deep learning applications, particularly in scenarios requiring collaborative inferencing across multiple devices for optimized real-time performance without cloud dependence\cite{b7}.
\subsection{Mutlimodal Models}
Multimodal models, particularly those designed for Visual Question Answering (VQA), process both images and user-posed questions to generate relevant responses. In our system, these models analyze the images captured by the smartphone and answer specific user questions about the environment, providing real-time, context-aware navigation assistance for visually impaired users. Research like the work on MiniGPT-4 \cite{b8} showcases how integrating advanced large language models with visual encoders can unlock sophisticated multi-modal abilities, enabling the model to generate detailed descriptions and context-aware answers based on visual inputs.

\subsection{Deep Learning and Neural Networks} 
Deep learning is a subcategory of machine learning, mainly based on the neural networks that model and solve complex problems by learning from enormous volumes of data. In particular, neural networks, of which the CNNs are a subset, have been great at recognizing images and performing other related types of analyses. Deep learning in our system has a very important place: it is crucial in signage detection and text extraction. It demonstrates how a pre-trained neural network can be applied in order to perform text extraction from images taken by smartphone cameras. When compared to the traditional image processing approaches, it outperforms them because of the ability to learn hierarchical features, such as edges and textures, directly from raw data. Current research, such as that presented by Shinde et al. (2023), points to the fact that the joint work of CNN and LSTM models can permit very accurate and speedy text extraction from document images \cite{b9}.

\subsection{Models Used}
In this study, four distinct methodologies were employed to detect and identify exit signs and other indoor signage. These approaches include the use of Multimodal Models, computer vision algorithms, and an API-based solution, each with unique advantages and limitations.

\begin{enumerate}
  \item \textbf{ViLT: }
  ViLT (Vision-and-Language Transformer) is a multimodal model that processes both image and text inputs \cite{b10}. For our implementation, we utilized a ViLT model pre-trained on the VQAv2 dataset. The model processes input images paired with specific queries, such as "Is this an exit sign?" or "Give a summary of the image," to generate relevant outputs. This model offers significant advantages, particularly its processing speed, which is crucial for real-time applications, and its multimodal capabilities, making it versatile for scenarios requiring simultaneous image and text interpretation. However, the model exhibits certain limitations in text recognition accuracy, such as misidentifying signage content. During our testing phase, for instance, the model incorrectly interpreted "Exit" signs as "Stop" signs. Such contextual misunderstandings significantly limit the model's practical application in complex navigation tasks.
  
  \item \textbf{BLIP Model: }{ The Bootstrapped Language-Image Pre-training (BLIP) provides another alternative with better visual-textual alignment \cite{b11}. This model was analyzed for its capability to increase the accuracy of text extraction from signage. BLIP showed better performance in text recognition compared to the ViLT, identifying the correct signage, for example, perceiving "Exit" when given with an Exit sign image. Its architecture is more robust under complicated visual situations; therefore, it stands out as more reliable in detecting signage. However, the large computation cost is the crucial drawback of this model, given that the model requires many resources in terms of its deployment and storage of parameters, which limits its feasibility for real-time applications on mobile devices.}
  \item \textbf{Astica.ai API: }
  {Astica.ai \cite{b12} offers an API-level service utilizing the most recent advancements in image recognition models for text extraction. It was integrated to enable the functionality of signage detection into smartphone applications. The API itself is very accurate and can be integrated without much overhead so that text can be interpreted efficiently without developing extensive in-house models such as Multimodal models and LLM's. Despite such advantages, the recurrent cost of 1 dollar per 1000 processed images brings a cost component, while depending on a third-party service does expose one to the risk of downtime and variations in terms of usage.}
  \item \textbf{EAST Text Detection Model: }
  {One of the text detection models applied in this work for signage detection is the EAST text detection model \cite{b13} model integrated with the OpenCV's DNN module. This is a text extraction solution that primarily relies on a pre-trained deep learning model for locating and extracting text from images. The EAST model is computationally efficient and highly accurate in detecting texts, even in the most complex environments; hence, it is suitable for deployment on standard hardware. This deep learning-based methodology yields robust results that are much more accurate than traditional image processing methods. However, the performance would be affected by differences in image quality, lighting conditions, camera angles, and noise, though to a lesser degree compared to traditional methods.}
  
 \end{enumerate}

\section{CHALLENGES AND RESEARCH OBJECTIVES}

Motivation:
Existing solutions for visually impaired people are often too complicated to use or require additional steps. For example, Microsoft's Seeing AI \cite{b14} app requires users to create waypoints initially, which the visually impaired can then follow later. Many existing solutions either involve multiple steps or necessitate special hardware to be carried by the user.
With the rise of AI and Large Language Models (LLMs), we now have the capability to infer and communicate information more clearly and precisely. The proposed solution addresses several key issues:

\begin{enumerate}
  \item \textbf{Privacy concerns that may arise from exposing building interiors to the internet.}
  \item \textbf{The challenge of energy consumption when running Multimodal models and LLMs on users' devices}
  \item \textbf{Compatibility with devices that lack advanced AI chips}
\end{enumerate}

\section{System Architecture}

The proposed system architecture implements a distributed computing approach, utilizing a Raspberry Pi-based processing unit in conjunction with a mobile application interface. This section details the system components, their interactions, and the underlying processing pipeline.

 \begin{figure}[h!]
\centering
\includegraphics[scale=0.25]{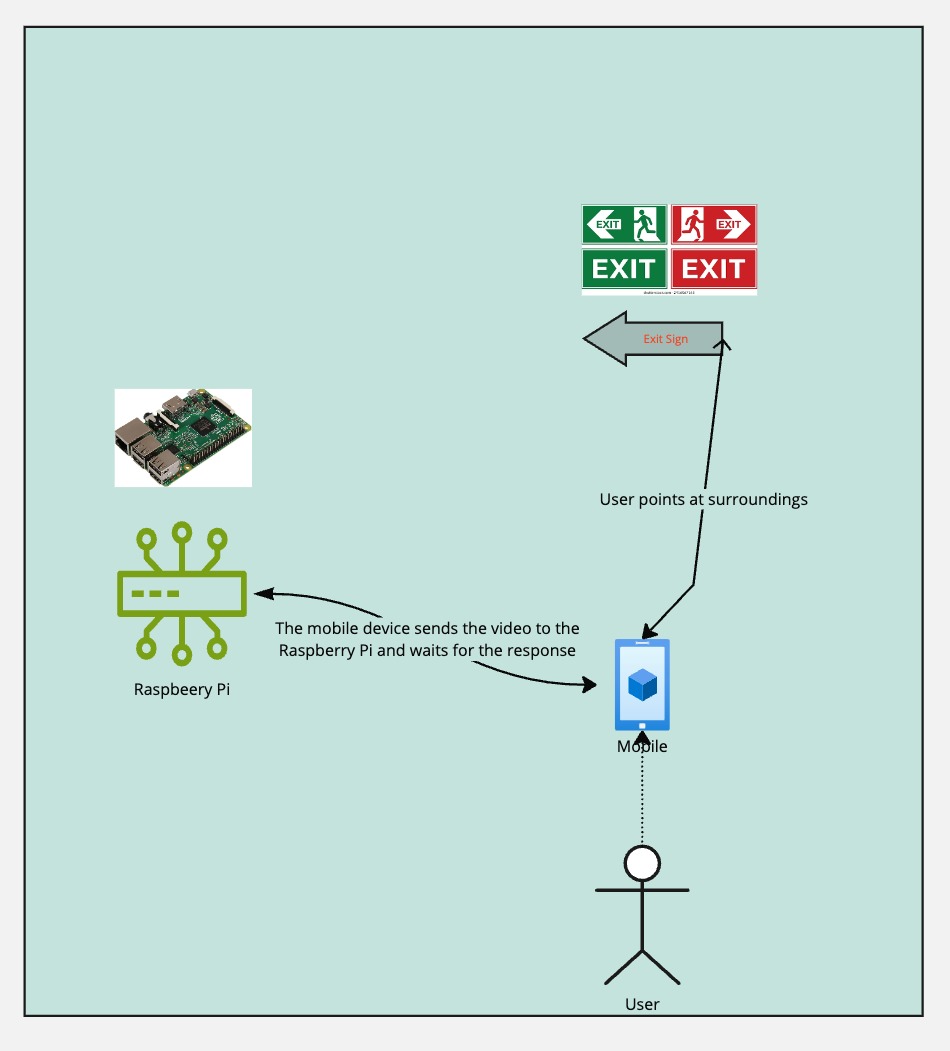}
\caption{System Architecture diagram showing the interaction between mobile device and Raspberry Pi processing unit}
\label{fig:method}
\end{figure}

\subsection{User Interaction Narrative:} 
When a visually impaired user interacts with the system, the process unfolds as follows:

\begin{enumerate}
\item \textbf{System Initiation}:
    \begin{itemize}
        \item User launches the mobile application
        \item Application establishes connection with nearby Raspberry Pi
        \item System confirms connection through audio feedback
    \end{itemize}

\item \textbf{Real-time Navigation}:
    \begin{itemize}
        \item User holds their smartphone, allowing the camera to capture the surroundings
        \item System continuously processes environmental data through the following sequence:
        \begin{enumerate}[]
            \item Camera captures real-time images/video
            \item Data is transmitted to Raspberry Pi
            \item Processing unit analyzes the scene
            \item Natural language instructions are generated
            \item Audio feedback is provided to the user
        \end{enumerate}
    \end{itemize}

\item \textbf{Environmental Interaction}:
    \begin{itemize}
        \item System identifies key elements such as:
        \begin{itemize}
            \item Doors and entrances
            \item Directional signs
            \item Potential obstacles
            \item Staircases and elevators
        \end{itemize}
        \item Provides context-aware instructions like:
        \begin{quote}
            ``There's an exit door 10 feet ahead on your right''\\
            ``Caution: stairs approaching in 5 steps''
        \end{quote}
    \end{itemize}

\item \textbf{Continuous Assistance}:
    \begin{itemize}
        \item System maintains constant monitoring and feedback
        \item Updates instructions based on user movement
        \item Responds to environmental changes in real-time
    \end{itemize}
\end{enumerate}

\subsection{Raspberry Pi System:}
The system is equipped with advanced multimodal and deep learning models capable of image recognition and text extraction. It is also installed with an on-device Large Language Model (LLM), such as Llama, which can convert simple text into detailed, descriptive narratives.

\subsection{Mobile Application:}
The mobile application serves as the user interface, establishing a connection with the Raspberry Pi system. It captures and transmits real-time video or images to the system, subsequently receiving detailed, turn-by-turn navigational instructions. This bidirectional communication facilitates a seamless interaction between the user's device and the processing unit, enabling efficient and accurate indoor navigation.

This solution offers several benefits:

\begin{enumerate}
  \item \textbf{Privacy Protection: } The Raspberry Pi, trained on local floor plan data and confined within the building, eliminates the risk of sensitive information being transmitted to external networks

\item \textbf{Energy Efficiency Analysis}
The system architecture demonstrates significant energy optimization through two primary mechanisms:

\subsubsection{Device-Level Optimization}
The distribution of computational load significantly reduces the energy consumption on user devices by:
\begin{itemize}
    \item Offloading compute-intensive tasks (model inference and video processing) to the Raspberry Pi
    \item Limiting mobile device operations to essential functions (image capture and audio playback)
    \item Minimizing wireless data transmission through localized processing
\end{itemize}

\subsubsection{System-Level Power Management}
The local processing architecture provides enhanced energy efficiency through:
\begin{itemize}
    \item Elimination of continuous cloud communication overhead
    \item Utilization of ARM-based processing architecture, optimized for low-power operation
    \item Implementation of hardware-accelerated model inference
\end{itemize}

  \item \textbf{Device Compatibility: }
The mobile application component enables interaction between various mobile devices (iOS or Android) and the multimodal, deep learning, and LLM models on the Raspberry Pi. This approach ensures broad accessibility, including for devices without advanced AI capabilities
 \end{enumerate}

\subsection{Internal working of Raspberry Pi}

\begin{figure}[h!]
\centering
\includegraphics[scale=0.25]{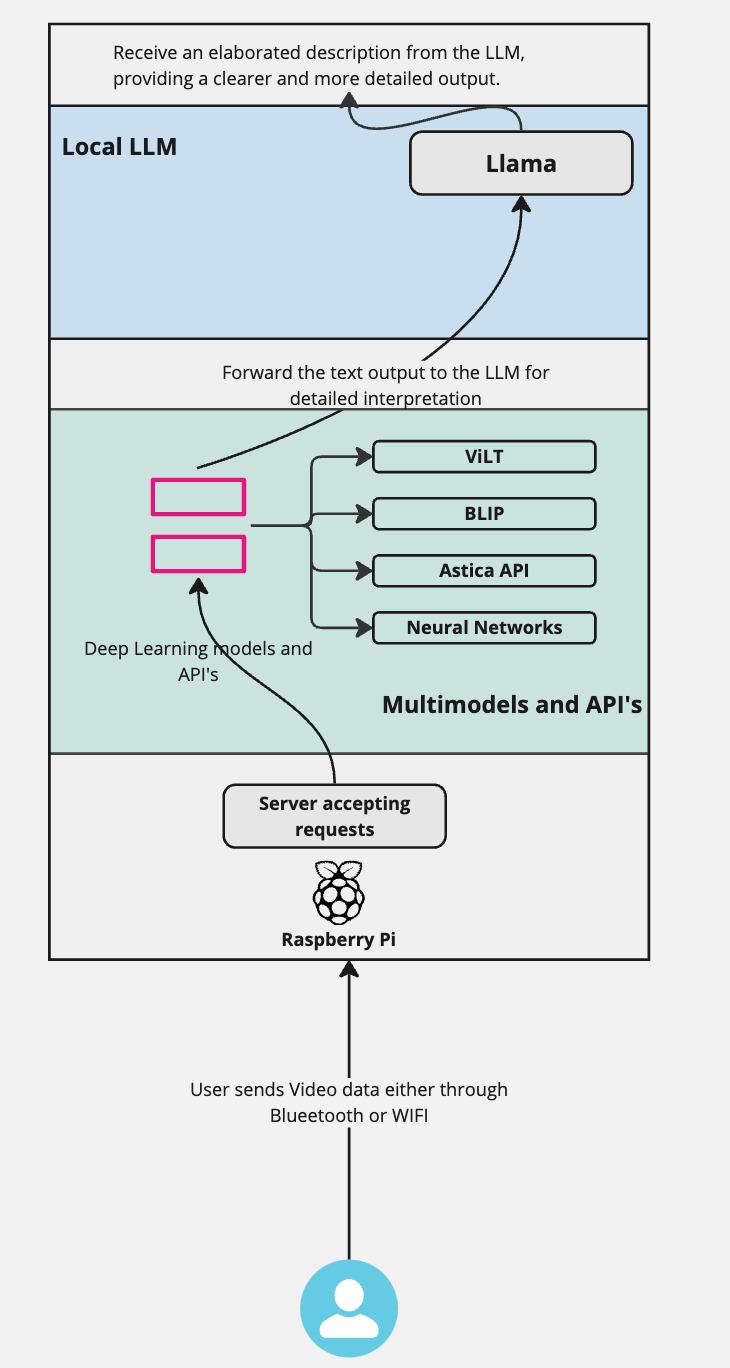}
\caption{Internals of Raspberry Pi}
\label{fig:method}
\end{figure}

The Raspberry Pi 5 is powered by a Broadcom ARM-based CPU with 8GB memory, offering a balance between efficiency and performance. In this system, the Raspberry Pi acts as the central processing unit for real-time image analysis and natural language instruction generation. When the smartphone captures images or videos, the data is transmitted via Wi-Fi to the Raspberry Pi. Upon receiving the visual data, the Pi processes it using multimodal models (BLIP or ViLT) or APIs. These models perform tasks such as object detection, signage recognition, and obstacle identification from the visual data. The identified information is then enhanced using large language models (LLMs) to generate context-aware natural language instructions. These instructions are sent back to the user’s smartphone as audio feedback, enabling seamless, real-time navigation. The Pi’s efficient design handles the computational workload, reducing strain on the user's mobile device while ensuring all data processing remains local for privacy.

\subsection{Evaluation and Key Metrics}
In reference to the paper \cite{b5}, where a comprehensive evaluation was conducted using the ISO 25010:2011 software quality standards, we intend to adopt a similar testing approach for our system. The evaluation will cover five key categories: Functional Stability, Performance Efficiency, Usability, Reliability, and Portability, ensuring that each aspect of our system is rigorously assessed. By utilizing the same testing techniques— such as measuring completeness, correctness, recoverability, and  learnability—we aim to ensure that our solution is robust, reliable, and  meets the specific needs of our users
\section{Conclusion and Future Work} 
The paper describes a successful demonstration of indoor navigation by a visually impaired user using a smartphone camera, Raspberry Pi, and modern AI models. The system utilizes deep learning, LLMs, and multimodal models to provide real-time, turn-by-turn audio guidance to the user based on the visual input while ensuring accessibility and ease of use.
As all the heavy processing is done locally with Raspberry Pi, the system will be lightweight on the user's device which enhances both the security and privacy of the user. This approach enables visually impaired individuals to navigate complex indoor layouts autonomously, this system strengthens their sense of independence and self-reliance, helping to break
down barriers that often restrict their access to daily life and shared space.

In future versions of the systems, other sensors, such as LiDAR or ultrasonic sensors, can be installed to provide better obstacle detection in low visibility or irregularly laid-out environments. This will definitely enhance the system's capability for navigation in more complex spaces. Future development of the system should also be directed at enabling it to pick up dynamic obstacles, such as people or other objects in movement, to further ensure its safe operation in an environment with unplanned activity. Further extension of multilingual support will make the system more available for a wider circle of users, whereas customized navigation instructions will allow users to tailor voice prompts to their personal preferences, making them more detailed and user-friendly.

\end{document}